\documentclass[runningheads]{llncs}
\usepackage{subcaption} 
\usepackage{multirow}
\usepackage{amsmath}
\usepackage{amssymb}
\usepackage{graphicx}
\usepackage{booktabs}
\usepackage{color}
\usepackage[colorlinks,
            linkcolor=red,
            anchorcolor=blue,
            citecolor=green
            ]{hyperref}
\usepackage{algorithmic}
\usepackage{algorithm}

\begin{document}
\title{A GPU Implementation of Multi-Guiding Spark Fireworks Algorithm for Efficient Black-Box Neural Network Optimization\thanks{Supported by organization x.}}
%
%
\author{
Xiangrui Meng\inst{1,2}\orcidID{0000-0002-2345-9485} \and
Ying Tan\inst{1,2,3,4}\orcidID{0000-0001-8243-4731}
}

\authorrunning{X. Meng and Y. Tan}
%
\institute{School of Intelligence Science and Technology, Peking University, Beijing, China \and
Key Laboratory of Machine Perception (MOE), Peking University, Beijing, China \and 
Institute for Artificial Intelligence, Peking University, Beijing, China\and
National Key Laboratory of General Artificial Intelligence, Peking University, Beijing, China \\
\email{mxxxr@stu.pku.edu.cn}, \email{ytan@pku.edu.cn}\\
}
\maketitle              
\begin{abstract}
Swarm intelligence optimization algorithms have gained significant attention due to their ability to solve complex optimization problems. However, the efficiency of optimization in large-scale problems limits the use of related methods. This paper presents a GPU-accelerated version of the Multi-Guiding Spark Fireworks Algorithm (MGFWA), which significantly improves the computational efficiency compared to its traditional CPU-based counterpart. We benchmark the GPU-MGFWA on several neural network black-box optimization problems and demonstrate its superior performance in terms of both speed and solution quality. By leveraging the parallel processing power of modern GPUs, the proposed GPU-MGFWA results in faster convergence and reduced computation time for large-scale optimization tasks. The proposed implementation offers a promising approach to accelerate swarm intelligence algorithms, making them more suitable for real-time applications and large-scale industrial problems. Source code is released at \textcolor{cyan}{https://github.com/mxxxr/MGFWA}.

\end{abstract}
\section{Introduction}

Black-box optimization is a critical process in many fields such as engineering, economics, and scientific research, where the objective is to find the best solution from a set of feasible solutions \cite{bbapplication}. Classical optimization methods, including gradient descent, linear programming, and dynamic programming, are effective for problems with well-defined mathematical models. However, these methods often struggle to handle real-world black-box problems that are complex, non-linear, or characterized by high-dimensional search spaces.

To address the limitations of traditional methods, nature-inspired optimization algorithms, commonly known as swarm intelligence algorithms, have been widely used. These algorithms simulate the collective behavior of natural systems, such as ant colony foraging \cite{ACO2006}, particle spreading \cite{PSO1995}, bee interaction \cite{BEE2009}, and collective fireworks explosion \cite{fwa2010}. Swarm intelligence algorithms have proven effective for solving complex optimization problems, particularly those that involve high-dimensional, multi-modal objective functions.

The Fireworks Algorithm (FWA) \cite{fwa2010} is a population-based optimization algorithm inspired by the explosion of fireworks. In FWA, each individual in the population represents a potential solution, and the algorithm simulates the explosion of fireworks. Each firework in the population emits sparks that search the solution space. The number and distance of these sparks are determined by the quality of the firework's corresponding solution, with higher-quality solutions generating more sparks that explore larger regions of the search space. The algorithm iteratively adjusts the position of the fireworks based on the feedback from the sparks, guiding the search toward better solutions.

FWA has gained attention due to its simplicity and its ability to effectively explore complex search spaces. However, like other swarm intelligence algorithms, it can suffer from issues such as premature convergence, where the algorithm gets stuck in local optima, or insufficient exploration of the search space. To address these limitations, the Multi-Guiding Spark Fireworks Algorithm (MGFWA) \cite{MGFWA2024} was proposed. MGFWA enhances FWA by generating more elite guiding sparks into the solution set, each of which performs more powerful exploitation ability using the proposed guiding mechanisms. This approach allows the algorithm to better exploitation the local search space, reduce the risk of premature convergence, and improve the overall convergence speed.

Despite the advantages of MGFWA, the computational cost remains a significant challenge. The computational expense arises primarily from the need to generate sparks and evaluate the objective function for each spark. This can become particularly prohibitive for large-scale optimization problems where the search space is high-dimensional and the objective function is complex. One promising solution to this problem is to leverage the parallel processing capabilities of modern Graphics Processing Units (GPUs). Unlike Central Processing Units (CPUs), which are optimized for serial processing, GPUs are designed to perform thousands of operations in parallel, making them ideal for computationally intensive tasks such as the evaluation of objective functions in optimization algorithms.

In this paper, we propose a GPU-accelerated version of the Multi-Guiding Spark Fireworks Algorithm (MGFWA), named GPU-MGFWA. This version enhances optimization efficiency by leveraging the parallel processing power of GPUs to simultaneously process multiple fireworks and sparks within the MGFWA framework. We conducted black-box neural network optimization experiments, and the results demonstrate that GPU-MGFWA outperforms MGFWA in terms of efficiency, while achieving the same optimization results. This approach can be adapted and extended to most variants of the fireworks algorithm.

\section{Related Work}

\subsection{Fireworks Algorithm and its Variants}

The Fireworks Algorithm (FWA) \cite{fwa2010} has undergone significant evolution and garnered increasing research attention due to its innovative mechanism inspired by fireworks explosions. Notable improvements include the enhanced FWA (EFWA) \cite{EFWA_2013}, which introduced adaptive control of explosion parameters, refined mutation mechanisms, and more efficient selection strategies, and the Dynamic Search FWA (dynFWA) \cite{zheng_dynamic_2014}, which adjusts explosion amplitudes dynamically based on search progress. Theoretical studies have also substantiated the algorithm’s convergence properties, global optimization capabilities, and adaptability, providing mathematical rigor to its performance improvements \cite{FWAAnalysis2014,FWAanalysis2017}. GFWA \cite{li_effect_2017} proposes elite bootstrap sparks that improve the performance of the fireworks algorithm for large-scale optimization problems. LoTFWA \cite{li_loser-out_2018} proposes a loser elimination mechanism that improves the exploration ability of the algorithm. SF-FWA \cite{chen2023sf} devised an adaptive mechanism for the algorithm's hyperparameter adaptive selection, as well as sampling distribution adaptation, and achieved good results in large-scale optimization. MGFWA \cite{MGFWA2024} proposes the multi-guiding spark mechanism, boosted guiding vector, and population-based random mapping to significantly improve the algorithm's exploitation ability.

Further extensions have tailored FWA to specific optimization paradigms, such as multi-objective optimization through techniques like S-metric-based selection \cite{liu2015s}, problem scalarization \cite{bejinariu2016fireworks}, and dynamic optimization using real-time adaptive strategies \cite{pekdemir2016enhancing}. Hybridizations with other algorithms, including Genetic Algorithms \cite{ye2017adaptive}, Particle Swarm Optimization \cite{chen2018ps}, and Differential Evolution \cite{zheng2015hybrid}, have leveraged complementary strengths to improve performance. The applications of FWA span a wide range of domains, including system security \cite{guo2019firework,he2013parameter}, image processing \cite{chen2017multilevel,rahmani2015privacy}, machine learning \cite{dutta2016artificial,gonsalves2016two}, and scheduling problems \cite{zheng2013multiobjective,ali2019optimising}, demonstrating its versatility. Despite its advancements, challenges such as handling constraints, scaling to large datasets, and improving interpretability remain active areas of research. These ongoing efforts indicate the algorithm’s potential for further development and broader applications in black-box optimization.

\subsection{GPU-Based Implementation of Swarm Intelligence Algorithms}

The parallel nature of swarm intelligence algorithms, where each agent’s computation is independent, makes them ideal candidates for GPU acceleration \cite{tan2015survey}. Early work in GPU-based swarm intelligence algorithms focused on parallelizing fitness evaluations and solution updates in Partical Swarm Optimization (PSO) \cite{GPUPSO2013}, Artificial Bee Colony (ABC) \cite{narasimhan2009parallel} and Ant Colony Optimization (ACO) \cite{GPUACO2009,GPUACO2013,GPUACO2014}. With the introduction of frameworks like CUDA \cite{CUDA} and OpenCL \cite{OPENCL}, researchers optimized memory management and thread synchronization, enabling more efficient GPU implementations \cite{merrill2011high}. In PSO, for example, the parallelization of both position and velocity updates, as well as fitness evaluations, allowed for larger populations and higher-dimensional problems to be handled more efficiently. The FWA also benefits from GPU acceleration. The parallelization of the explosion and mutation operators in FWA allows for concurrent processing of large numbers of sparks, greatly enhancing the algorithm's efficiency \cite{ding2013gpu}. 

Hybrid systems combining GPUs and CPUs have further advanced swarm intelligence algorithms \cite{catala2007strategies,GPUACO2009}. GPUs handle the parallelizable components, such as fitness evaluations, while CPUs manage sequential tasks like agent interactions and data structure maintenance \cite{fu2010parallel,cecilia2011parallelization}. However, challenges such as memory management in large populations and optimizing thread load balancing remain. Techniques like efficient memory hierarchies \cite{dawson2013improving}, branch divergence \cite{narasimhan2009parallel} and reducing random numbers \cite{uchida2012efficient} have been proposed to address these issues. The GPU-based swarm intelligence algorithms has been applied to many complex real-world problems, such as energy consumption \cite{iruela2024gpu}, medicine \cite{abbas2024artificial}, economy \cite{tewatia2023gpu} and machine learning \cite{chaudhari2024psogsa}.

\section{Implementation of GPU-MGFWA}

The MGFWA, which combines the strengths of the FWA and the multi-guiding spark mechanism, has already demonstrated promising results in terms of solution quality and exploration capabilities. However, its computational cost can still be a limiting factor. 

Using GPUs to accelerate the computation can significantly reduce the time required for each iteration, especially when evaluating high-dimensional objective functions. As a swarm intelligence algorithm, the independent evaluations of fireworks and sparks in MGFWA can be performed in parallel on the GPU, leading to a substantial reduction in computation time. GPU acceleration enables the efficient handling of large-scale optimization problems in MGFWA, where the high-dimensional search spaces would otherwise make the algorithm prohibitively slow.

\begin{table}[t]
\centering
\caption{Variable setting.}
\begin{tabular}{ccccl} \toprule
Alg                        & Variable & Clarification   & Shape &                      \\ \midrule
\multirow{4}{*}{MGFWA}     & $F$      & Firework        & $N \times D$   &                      \\
                           & $S_e$    & Explosion Spark & $N \times D$   &                      \\
                           & $S_g$    & Guiding Spark   & $N \times D$   &                      \\
                           & $Y$      & Fitness         & $N \times D$   &                      \\ \midrule
\multirow{4}{*}{GPU-MGFWA} & $F$      & Firework        & $B \times N \times D$ &                      \\
                           & $S_e$    & Explosion Spark & $B \times N \times D$ & \multicolumn{1}{c}{} \\
                           & $S_g$    & Guiding Spark   & $B \times N \times D$ & \multicolumn{1}{c}{} \\
                           & $Y$      & Fitness         & $B \times N \times D$ & \multicolumn{1}{c}{} \\ \bottomrule
\end{tabular}
\label{tab:variable}
\end{table}

To implement the GPU-accelerated version of MGFWA, we utilize PyTorch \cite{pytorch}, a popular open-source deep learning framework known for its flexibility and performance. PyTorch provides an intuitive interface for tensor operations and seamless integration with both CPUs and GPUs. PyTorch's support for CUDA, a parallel computing platform and API model developed by NVIDIA, allows us to harness the power of GPUs for accelerating computations. By using PyTorch's GPU functionality, we can efficiently parallelize the evaluation of fireworks and sparks, significantly speeding up the algorithm’s convergence process. Moreover, PyTorch's optimization capabilities and built-in functions for tensor manipulation simplify the development of custom optimization algorithms, making it an excellent choice for implementing GPU-MGFWA.

\begin{figure}[t]
\centerline{\includegraphics[width=\textwidth]{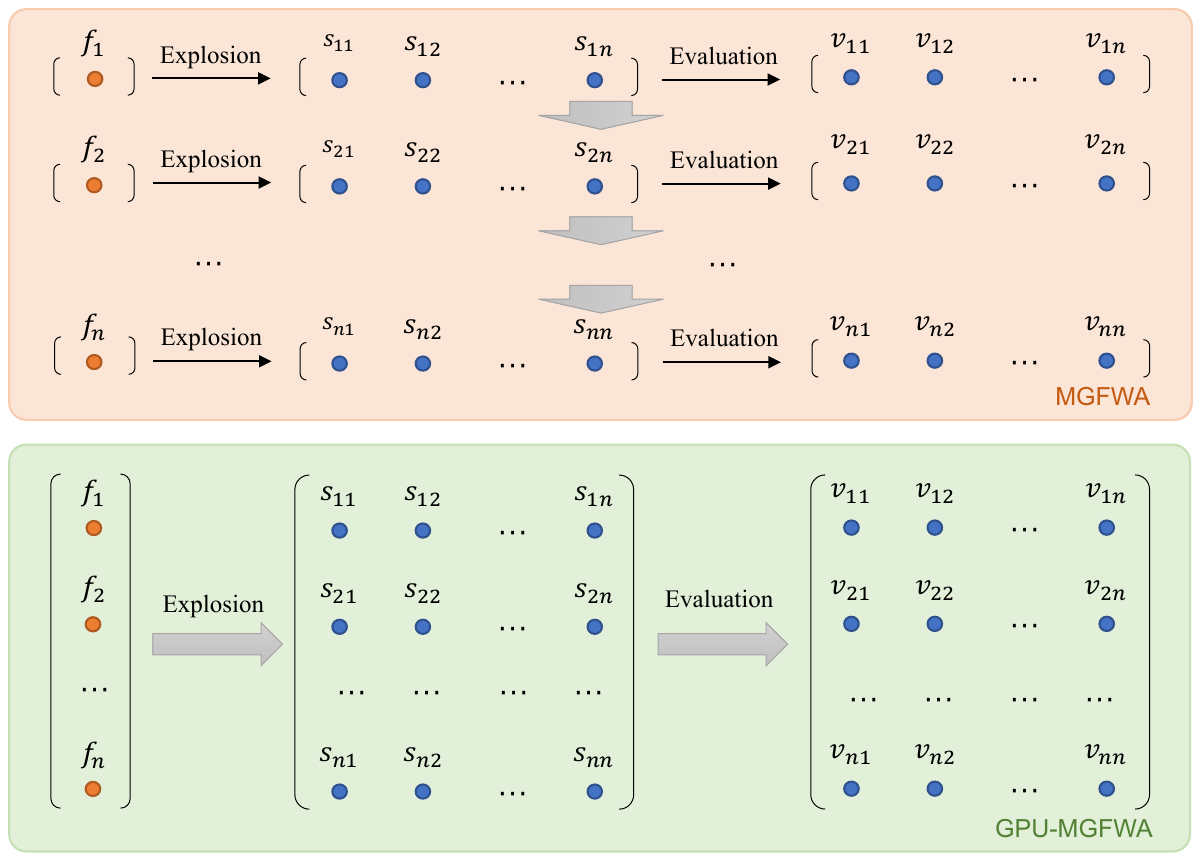}}
\caption{Comparison of the two algorithms for explosion operation. MGFWA: sparks are generated and evaluated for each firework, all fireworks are processed serially. GPU-MGFWA: sparks are generated and evaluated for all fireworks at the same time, all fireworks are processed in parallel.}
\label{fig:comp}
\end{figure}

To leverage the parallel computing power of the GPU, we introduced a batch dimension to all variables in the algorithm, enabling the evaluation of all fireworks and sparks across multiple batches during each evaluation. The definitions of the relevant variables are provided in Table \ref{tab:variable}. Additionally, we modified the MGFWA, replacing the serial computation and evaluation of each generation of fireworks with parallel processing. Through these enhancements, including multi-batch parallel evaluation and multi-firework parallel computation, we have significantly improved the parallel efficiency of the algorithm while preserving the optimization capabilities of the MGFWA. The flow of the GPU-MGFWA is shown in Algorithm \ref{alg:1}. The comparison of the explosion process between the two algorithms is shown as an example in Fig. \ref{fig:comp}. 

\begin{algorithm}[t]
\caption{The process of GPU-MGFWA}
\label{alg:1}
\begin{algorithmic}[1]
\STATE \textbf{Input}: Target black-box to optimize $f$.
\STATE \textbf{Output}: The best fitness of $f$ and its corresponding firework position.
\STATE Randomly initialize $\mu$ fireworks of $b$ batches in the potential space.
\REPEAT
    \STATE \textbf{Parallel Generate explosion sparks of MGFWA by GPU.}
    \STATE Perform the population-based random mapping to explosion sparks.
    \STATE \textbf{Parallel Generate multi-guiding spark of MGFWA by GPU.}
    \STATE Perform the population-based random mapping to guiding sparks.
    \STATE \textbf{Parallel evaluate all the fitness of the sparks by GPU.}
    \STATE Select the best individual of each population.
    \STATE Perform the loser-out tournament of MGFWA.
\UNTIL{termination criteria is met.}
\RETURN the position and the fitness of the best individual.
\end{algorithmic}
\label{alg_MGFWA}
\end{algorithm}

\section{Experiments}

\subsection{Testbeds and Experimental Setup}
\subsubsection{Testbeds:} We evaluated the algorithms using neural networks as black-boxes for optimization. Specifically, the parameters of the neural network are fixed, and the goal is to find the input that minimizes the network's output. Assume that the input to the neural network has $d$ dimensions and the solution space to be optimized is $\mathbb{R}^{d}$, we consider the following minimization problem:
\begin{equation}
\min _{\mathbf{x} \in \mathbb{R}^{d}} f(\mathbf{x}) \label{eq}
\end{equation}

where $\mathbf{x}$ is a vector in the solution space, and $f$ represents the black-box neural network with fixed parameters. The output of $f(\mathbf{x})$ which is also called fitness is the object needed to minimize. We designed three different scales of neural networks—small, medium, and large—and the network configurations are shown in Table \ref{tab:nnsettings}.

\begin{table}[t]
\centering
\caption{Neural network configurations used for black-box optimization.}
\begin{tabular}{cccccccc}
\toprule
Net ID & Scale                   & Activation & Input dim             & Hidden dim & Output dim         & Layer num           & Params. \\ \hline
1      & \multirow{4}{*}{Small}  & ReLU       & \multirow{2}{*}{10}   & 16         & \multirow{4}{*}{1} & \multirow{4}{*}{4}  & 465     \\
2      &                         & GELU       &                       & 32         &                    &                     & 4,609    \\ \cline{4-4}
3      &                         & ReLU       & \multirow{2}{*}{20}   & 16         &                    &                     & 1,441    \\
4      &                         & GELU       &                       & 32         &                    &                     & 4,929    \\ \hline
5      & \multirow{4}{*}{Medium} & ReLU       & \multirow{2}{*}{100}  & 64         & \multirow{4}{*}{1} & \multirow{4}{*}{7}  & 35,649   \\
6      &                         & GELU       &                       & 128        &                    &                     & 128,641  \\ \cline{4-4}
7      &                         & ReLU       & \multirow{2}{*}{200}  & 64         &                    &                     & 42,049   \\
8      &                         & GELU       &                       & 128        &                    &                     & 141,441
  \\ \hline
9      & \multirow{4}{*}{Large}  & ReLU       & \multirow{2}{*}{1000} & 256        & \multirow{4}{*}{1} & \multirow{4}{*}{10} & 914,433  \\
10     &                         & GELU       &                       & 512        &                    &                     & 3,137,585 \\ \cline{4-4}
11     &                         & ReLU       & \multirow{2}{*}{2000} & 256        &                    &                     & 1,173,433 \\
12     &                         & GELU       &                       & 512        &                    &                     & 3,657,585 \\ \bottomrule
\end{tabular}
\label{tab:nnsettings} 
\end{table}

\subsubsection{Experimental Setup:} Experiments are performed on a computing server with 256 GB RAM, one 64-core CPU, and one GeForce RTX 3090 GPU used for sampling and evaluation. We use Pytorch 2.4 and CUDA 11.8 in our experiments. More information of software version can be found at our source code. 

\subsection{Main Results}

\begin{figure}[htbp] 
    \centering 
    \begin{subfigure}{0.38\textwidth}
        \centering
        \includegraphics[width=\textwidth]{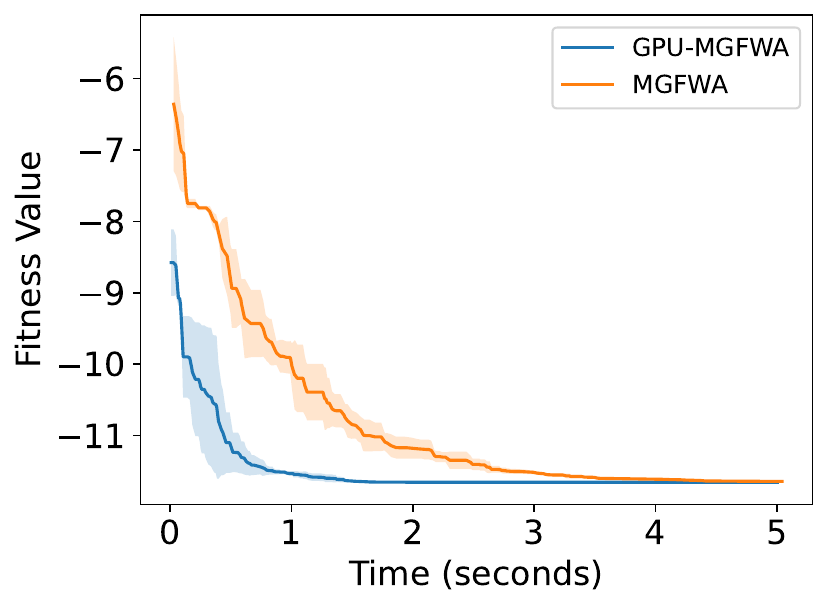}
        \caption{Net 1} 
        \label{fig:image1} 
    \end{subfigure}
    \hfill 
    \begin{subfigure}{0.38\textwidth}
        \centering
        \includegraphics[width=\textwidth]{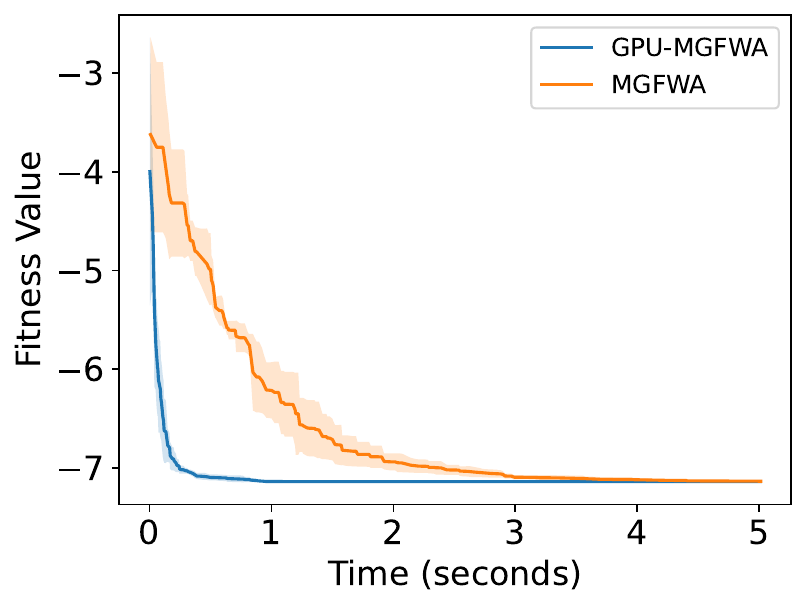}
        \caption{Net 2}
        \label{fig:image2}
    \end{subfigure}
    \begin{subfigure}{0.38\textwidth}
        \centering
        \includegraphics[width=\textwidth]{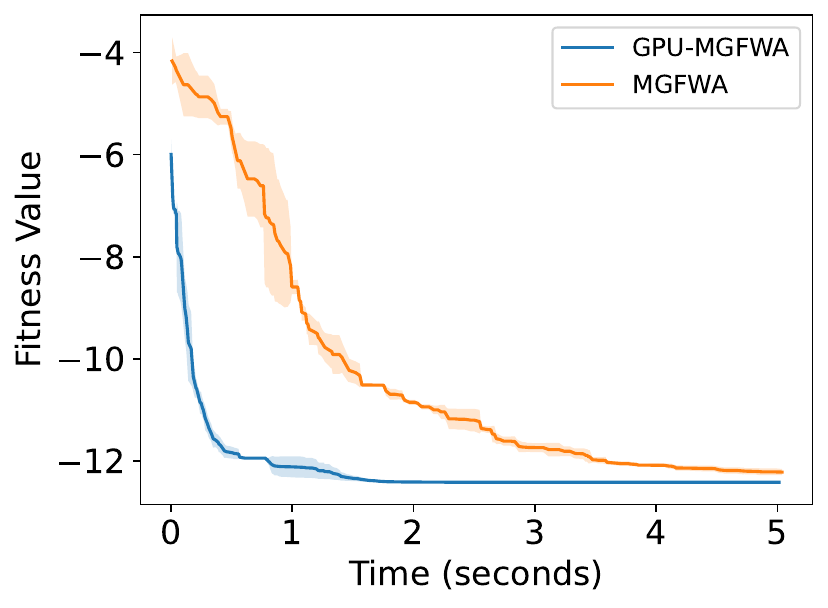}
        \caption{Net 3}
        \label{fig:image4}
    \end{subfigure}
    \hfill
    \begin{subfigure}{0.38\textwidth}
        \centering
        \includegraphics[width=\textwidth]{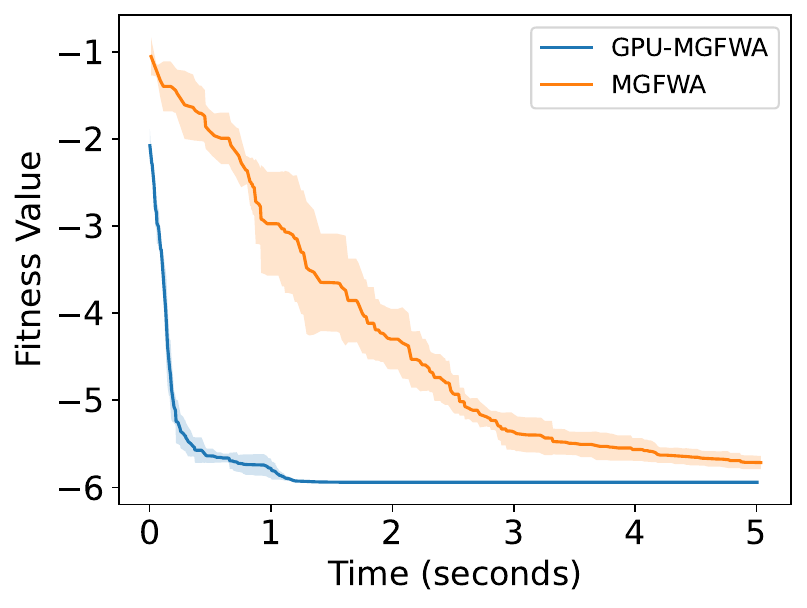}
        \caption{Net 4}
        \label{fig:image5}
    \end{subfigure}
    \caption{Optimization Efficiency Comparison between MGFWA and GPU-MGFWA on \textit{small} scale Black-Box Neural Network Optimization} 
    \label{fig:smallresults} 
\end{figure}

\begin{figure}[htbp] 
    \centering 
    \begin{subfigure}{0.38\textwidth} 
        \centering
        \includegraphics[width=\textwidth]{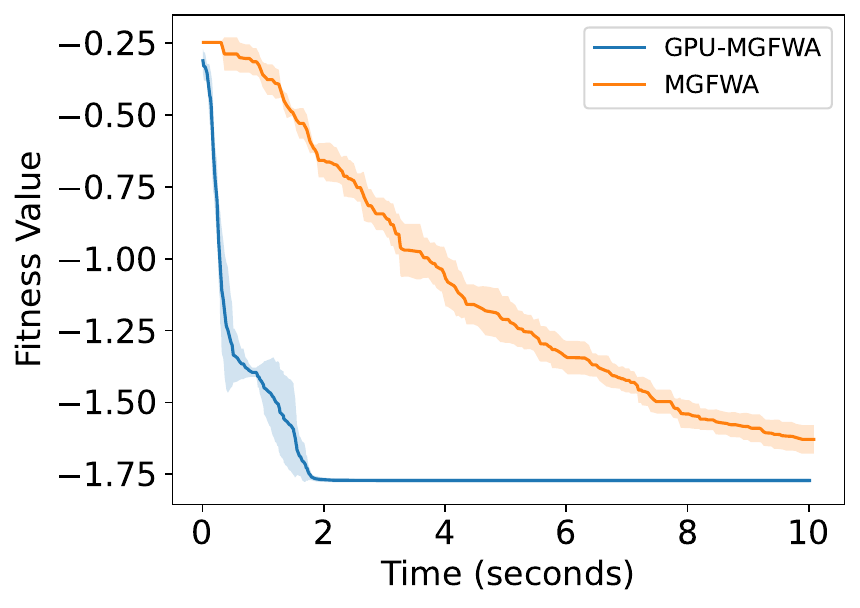} 
        \caption{Net 5} 
        \label{fig:image1} 
    \end{subfigure}
    \hfill
    \begin{subfigure}{0.38\textwidth}
        \centering
        \includegraphics[width=\textwidth]{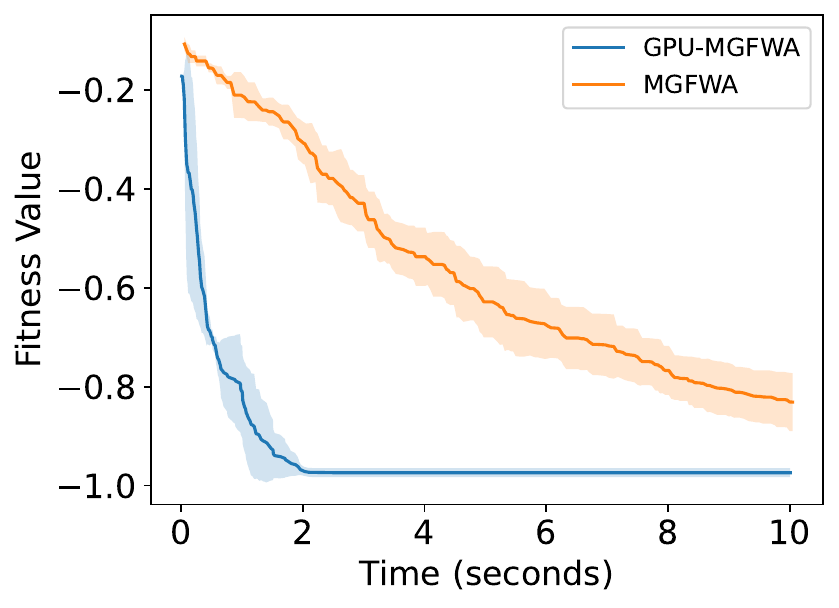}
        \caption{Net 6}
        \label{fig:image2}
    \end{subfigure}
    \begin{subfigure}{0.38\textwidth}
        \centering
        \includegraphics[width=\textwidth]{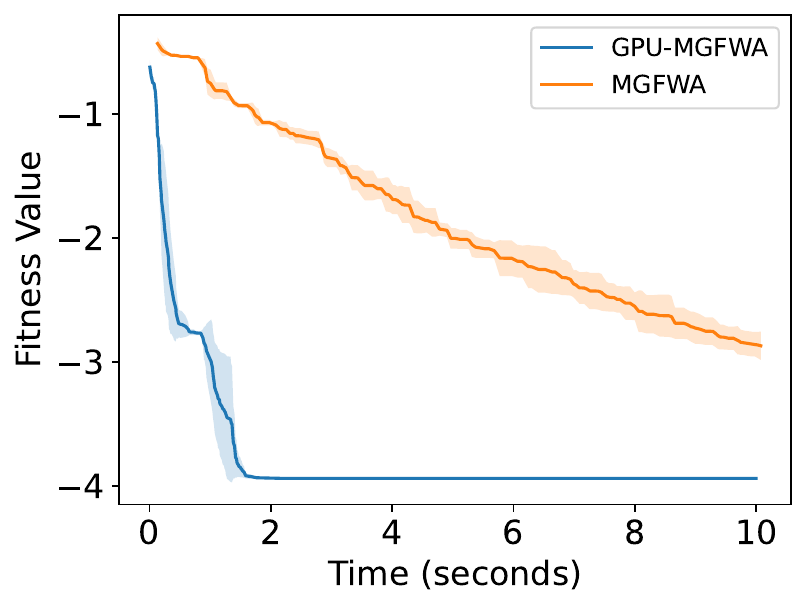}
        \caption{Net 7}
        \label{fig:image4}
    \end{subfigure}
    \hfill
    \begin{subfigure}{0.38\textwidth}
        \centering
        \includegraphics[width=\textwidth]{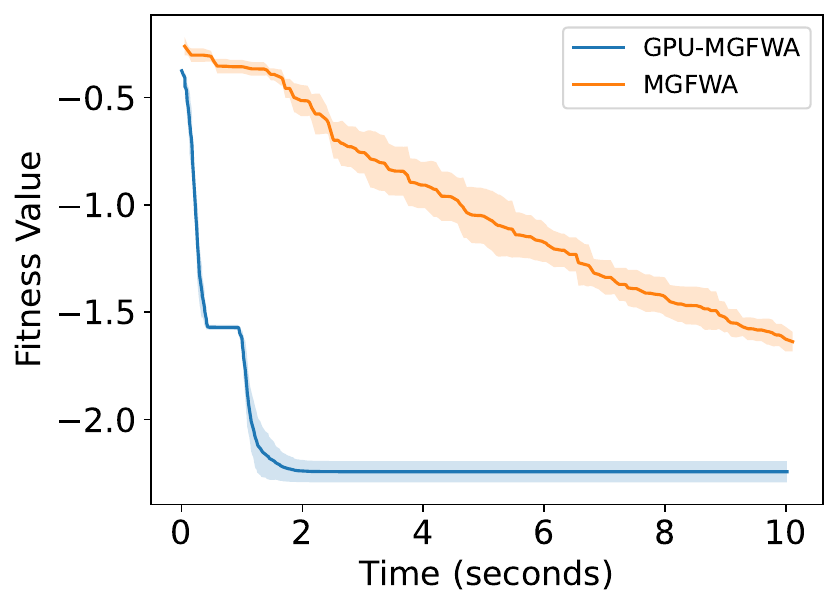}
        \caption{Net 8}
        \label{fig:image5}
    \end{subfigure}
    \caption{Optimization Efficiency Comparison between MGFWA and GPU-MGFWA on \textit{medium} scale Black-Box Neural Network Optimization}
    \label{fig:mediumresults} 
\end{figure}

We generated 12 neural networks based on the settings in Table \ref{tab:nnsettings}, with their parameters randomly sampled and fixed for each network. The optimization results, aimed at minimizing the output of each network, are presented in Figures \ref{fig:smallresults}, \ref{fig:mediumresults}, and \ref{fig:largeresults}. Shaded regions indicate standard deviations across 8 runs. It can be observed that GPU-MGFWA, leveraging GPU parallel acceleration, consistently outperforms MGFWA across all given timeframes. Moreover, the efficiency advantage of GPU-MGFWA becomes increasingly pronounced as the network size grows.

For smaller neural networks, shorter optimization times are designed, while larger networks are allocated longer optimization periods. The results clearly demonstrate that GPU-MGFWA achieves significantly faster convergence than MGFWA, as reflected in the steeper descent of the blue curves compared to the orange ones. GPU-MGFWA not only reaches lower fitness values in a shorter amount of time but also exhibits greater consistency, as evidenced by the narrower shaded regions around its curves. For small-scale optimization scenarios (Fig. \ref{fig:smallresults}), GPU-MGFWA converges rapidly within one second, whereas MGFWA requires considerably more time to achieve comparable results. In more complex scenarios (Fig. \ref{fig:mediumresults}), GPU-MGFWA maintains its advantage, converging faster and achieving similar or better fitness values. For large-scale optimization scenarios (Fig. \ref{fig:largeresults}), MGFWA struggles to optimize efficiently within the given timeframe, whereas GPU-MGFWA continues to demonstrate efficient optimization even at larger scales and within a shorter duration (within 5 seconds). These findings highlight the computational benefits of GPU acceleration, particularly in handling high-dimensional or complex optimization problems, establishing GPU-MGFWA as a highly efficient solution for large-scale and time-sensitive applications.

\begin{figure}[htbp] 
    \centering 
    \begin{subfigure}{0.38\textwidth}
        \centering
        \includegraphics[width=\textwidth]{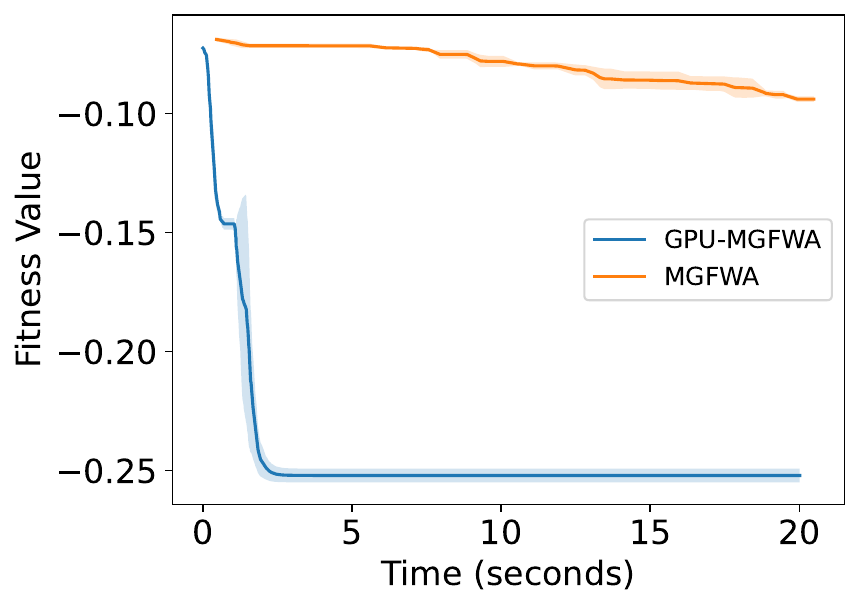} 
        \caption{Net 9} 
        \label{fig:image1}
    \end{subfigure}
    \hfill 
    \begin{subfigure}{0.38\textwidth}
        \centering
        \includegraphics[width=\textwidth]{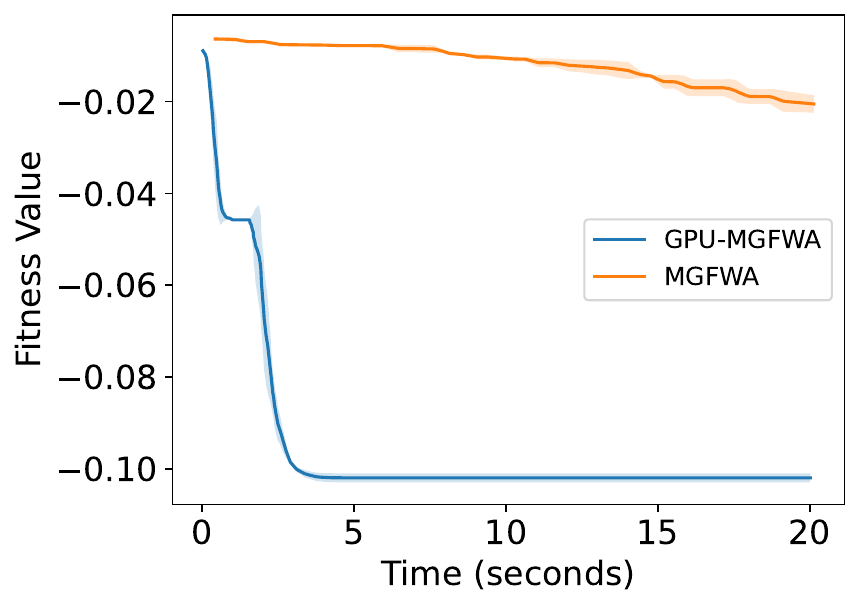}
        \caption{Net 10}
        \label{fig:largeresults}
    \end{subfigure}
    \begin{subfigure}{0.38\textwidth}
        \centering
        \includegraphics[width=\textwidth]{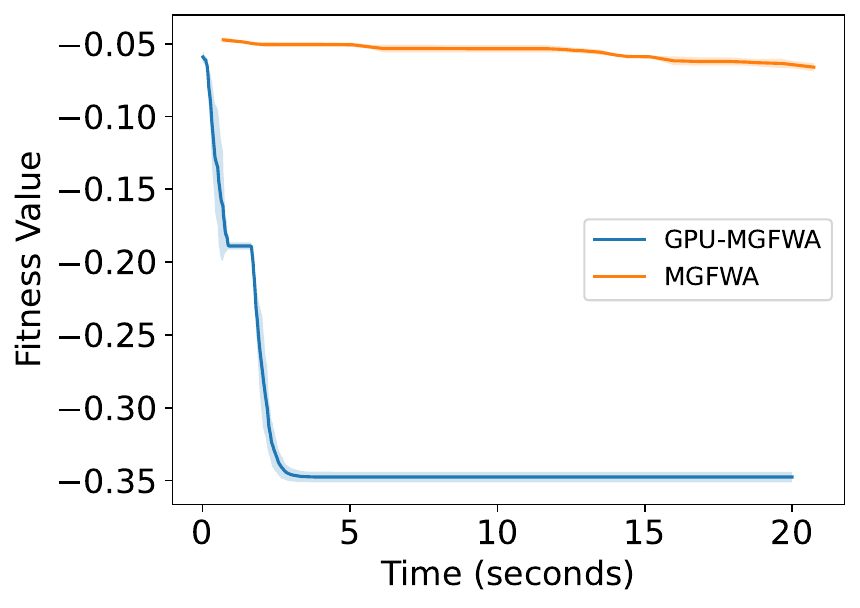}
        \caption{Net 11}
        \label{fig:image4}
    \end{subfigure}
    \hfill
    \begin{subfigure}{0.38\textwidth}
        \centering
        \includegraphics[width=\textwidth]{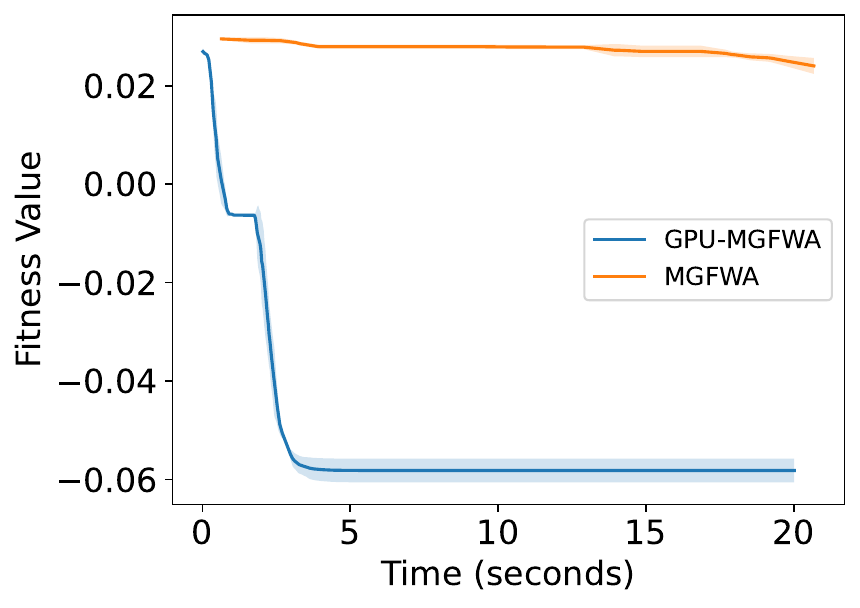}
        \caption{Net 12}
        \label{fig:image5}
    \end{subfigure}
    \caption{Optimization Efficiency Comparison between MGFWA and GPU-MGFWA on \textit{large} scale Black-Box Neural Network Optimization}
    \label{fig:largeresults} 
\end{figure}

\section{Conclusion}
In this paper, we present a GPU-accelerated version of the Multi-Guiding Spark Fireworks Algorithm (MGFWA), to significantly improve the computational efficiency of FWA. By taking advantage of the parallel processing capabilities of modern GPUs, the GPU-MGFWA drastically reduces the computational time, making it more suitable for real-time and large-scale applications. The results of experiments on neural network black-box optimization problems show that the GPU-MGFWA performs not only faster but also maintains or even improves the quality of the solutions compared to the CPU-based version. The speedup achieved through GPU acceleration opens new possibilities for applying MGFWA to more complex, real-world optimization problems that require faster convergence times and greater computational efficiency.

While our GPU-MGFWA demonstrates significant advancements in optimization efficiency, there are still avenues for further research and enhancement. Future work could explore more advanced parallelization techniques, hybrid optimization strategies that combine MGFWA with other optimization methods, and the application of the algorithm to more diverse and complex real-world problems, such as those found in machine learning, engineering design, and data science.

\subsubsection{Acknowledgments.} This work is supported by the National Natural Science Foundation of China (Grant No. 62250037, 62076010 and 62276008), and partially supported by Science and Technology Innovation 2030-“New Generation Artificial Intelligence” Major Project (Grant Nos.: 2018AAA0102301).

\bibliographystyle{splncs04}
\bibliography{mgfwa}

\end{document}